\documentclass[conference]{IEEEtran}
\usepackage{graphicx}
\usepackage{amsmath}
\usepackage{amssymb}
\usepackage{booktabs}
\usepackage{multirow}
\usepackage{url}

\usepackage{hyperref}
\hypersetup{
    pdfborder={0 0 0} 
}

\usepackage{orcidlink}
\usepackage{fancyhdr} 

\fancypagestyle{firstpageheader}{%
  \fancyhf{} 

  \fancyhead[C]{\fontfamily{ptm}\fontsize{10}{12}\selectfont \textnormal{2025 12th International Conference on Emerging Trends in Engineering \& Technology - Signal and Information Processing (ICETET - SIP)}}

  \fancyfoot[L]{%
    \parbox[t]{\columnwidth}{%
      \footnotesize
      \textnormal{979-8-3315-0099-3/25/\$31.00~\copyright~2025 IEEE}%
      \vspace*{1pt}
    }%
  }
}

\fancypagestyle{subsequentpages}{%
  \fancyhf{} 
  
  \fancyhead[C]{\fontfamily{ptm}\fontsize{10}{12}\selectfont \textnormal{2025 12th International Conference on Emerging Trends in Engineering \& Technology - Signal and Information Processing (ICETET - SIP)}}
  
  \fancyfoot[C]{\thepage}
}

\IEEEoverridecommandlockouts
\IEEEpubid{\makebox[\columnwidth][l]{\textnormal{979-8-3315-0099-3/25/\$31.00~\copyright~2025 IEEE}\hfil} \hspace{\columnsep}\makebox[\columnwidth][l]{ }}

\begin{document}

\title{A Novel AI-Driven System for Real-Time Detection of Mirror Absence, Helmet Non-Compliance, and License Plates Using YOLOv8 and OCR}

\author{
    \IEEEauthorblockN{Nishant V H\orcidlink{0009-0009-8163-4489}}
    \IEEEauthorblockA{\textit{Computer Science and Engineering} \\
    \textit{RV College of Engineering} \\
    Bengaluru, India \\
    nishantvh.cy23@rvce.edu.in}
\and
    \IEEEauthorblockN{Aditi Agarwal}
    \IEEEauthorblockA{\textit{Computer Science and Engineering} \\
    \textit{RV College of Engineering} \\
    Bengaluru, India \\
    aditiagarwal.cy23@rvce.edu.in}
\and
    \IEEEauthorblockN{Minal Moharir\orcidlink{0000-0001-8256-5440}}
    \IEEEauthorblockA{\textit{Computer Science and Engineering} \\
    \textit{RV College of Engineering} \\
    Bengaluru, India \\
    minalmoharir@rvce.edu.in}
}

\maketitle
\thispagestyle{firstpageheader}
\pagestyle{subsequentpages} 
\begin{abstract}
Road safety is a critical global concern, with manual enforcement of helmet laws and vehicle safety standards (e.g., rear-view mirror presence) being resource-intensive and inconsistent. This paper presents an AI-powered system to automate traffic violation detection, significantly enhancing enforcement efficiency and road safety. The system leverages YOLOv8 for robust object detection and EasyOCR for license plate recognition. Trained on a custom dataset of annotated images (augmented for diversity), it identifies helmet non-compliance, the absence of rear-view mirrors on motorcycles—an innovative contribution to automated checks—and extracts vehicle registration numbers. A Streamlit-based interface facilitates real-time monitoring and violation logging. Advanced image preprocessing enhances license plate recognition, particularly under challenging conditions. Based on evaluation results, the model achieves an overall precision of \textbf{0.9147}, recall of \textbf{0.886}, and a mean Average Precision (mAP@50) of \textbf{0.843}. The mAP@50–95 of \textbf{0.503} further indicates strong detection capability under stricter IoU thresholds. This work demonstrates a practical and effective solution for automated traffic rule enforcement, with considerations for real-world deployment discussed.
\end{abstract}

\begin{IEEEkeywords}
Two-wheeler safety, YOLOv8, helmet violation detection, rear-view mirror detection, license plate recognition, EasyOCR, Roboflow, Streamlit interface, Computer Vision, Traffic monitoring system.
\end{IEEEkeywords}

\section{Introduction}
\label{sec:introduction}
Road safety is a paramount global concern, with traffic accidents leading to a significant number of fatalities and injuries annually \cite{who_road_safety}. Despite the implementation of stringent traffic regulations, non-compliance remains prevalent. Traditional manual enforcement methods are often labor-intensive, costly, and susceptible to human inconsistency, underscoring the urgent need for more efficient, scalable, and objective solutions. The automation of traffic rule enforcement, as demonstrated in recent intelligent frameworks \cite{hegde2024intelligent}, can play a pivotal role in mitigating these challenges, leading to improved adherence to safety standards and, consequently, a reduction in road incidents.

This paper presents a novel AI-driven system designed for the real-time, automated detection of multiple traffic violations. Specifically, the system leverages the state-of-the-art YOLOv8 object detection model—a technology proven effective for key tasks like helmet detection \cite{Patel2023ICPCSN}—to identify (1) motorcyclists not wearing helmets and (2) motorcycles operating without essential rear-view mirrors—a critical safety aspect often overlooked in automated systems. Furthermore, the system integrates EasyOCR for accurate extraction of vehicle registration numbers from license plates, facilitating violator identification. A user-friendly interface, developed using Streamlit, enables real-time monitoring, visualization of detected violations, and efficient logging of incidents.

By automating these detection and identification processes, our proposed system aims to:
\begin{itemize}
    \item Reduce the operational burden on law enforcement personnel.
    \item Enhance the consistency and reach of traffic rule enforcement.
    \item Improve compliance with crucial safety regulations.
    \item Ultimately contribute to safer road environments and a reduction in traffic-related casualties.
\end{itemize}
This work not only demonstrates the technical feasibility of such a comprehensive system but also highlights its practical applicability in real-world traffic management scenarios.

The remainder of this paper is organized as follows: Section \ref{sec:background} reviews related work in automated traffic violation detection. Section \ref{sec:methodology} details the proposed system architecture, dataset creation, preprocessing techniques, and violation detection logic. Section \ref{sec:results_discussion} presents the experimental results and a thorough discussion of the system's performance. Finally, Section \ref{sec:conclusion} concludes the paper, summarizing key findings and outlining directions for future research, including practical deployment considerations and ethical implications.

\section{Background Study and Related Work}
\label{sec:background}
Automated traffic rule enforcement has become a critical area of research due to the limitations of manual methods, which are often labor-intensive, prone to human error, and inconsistent. Advances in artificial intelligence (AI) and computer vision offer promising alternatives for enhancing detection accuracy and enabling real-time monitoring, crucial for timely intervention in dynamic traffic scenarios.

Early research in this domain, such as Kulkarni et al. \cite{Kulkarni2018} and Kadam et al. \cite{Kadam2021}, focused on integrating Automatic Number Plate Recognition (ANPR) with helmet detection. These studies utilized traditional Convolutional Neural Network (CNN) based approaches for feature extraction and object detection. While foundational, these systems faced significant limitations: OCR accuracy was often compromised in challenging conditions (e.g., low-light, high noise), and scalability to diverse real-world scenarios was restricted. For instance, the work by Darji et al. \cite{Darji2020} explored lightweight CNNs for license plate identification; however, its performance was primarily validated in controlled environments, failing to fully address complexities introduced by variations in weather, lighting, or partial occlusions. These earlier systems typically struggled with maintaining high accuracy and real-time processing speeds simultaneously, often requiring a trade-off between the two.

More recent advancements have seen the adoption of sophisticated deep learning models, particularly from the You Only Look Once (YOLO) family, known for their superior balance of speed and accuracy. Patel et al. \cite{Patel2023ICPCSN} demonstrated the viability of YOLOv8 for helmet detection in industrial settings, but its application was confined to worker safety and did not address the multifaceted nature of traffic monitoring. Similarly, Vijendar Reddy et al. \cite{VijendarReddy2024} reviewed approaches including YOLOv8 with EasyOCR for concurrent helmet and license plate detection. However, their focus, like many others, remained on a limited set of violations, often neglecting other critical safety features. Aquitan et al. \cite{Aquitan2024} emphasized the real-time processing benefits of YOLOv8 for helmet detection but did not extend their system to incorporate multiple violation types or crucial elements like rear-view mirror checks. A common thread in these recent works is the incremental adoption of newer YOLO versions for specific tasks, but a comprehensive, integrated system addressing multiple, diverse two-wheeler safety violations, particularly in contexts like Indian roads where real-time traffic management solutions are crucial \cite{raman2020hybrid}, has been less explored.

While these studies highlight progress, several gaps persist:
\begin{itemize}
    \item \textbf{Limited Scope of Violations:} Most existing systems focus on one or two violation types (primarily helmet and ANPR), neglecting other important safety aspects like the mandatory presence of rear-view mirrors on motorcycles.
    \item \textbf{Preprocessing for OCR:} While some works mention ANPR, the impact of specific, advanced preprocessing pipelines tailored for enhancing OCR robustness under varied conditions is often not deeply investigated or compared.
    \item \textbf{Real-world Robustness:} Many systems are validated on limited datasets or under constrained conditions, with insufficient analysis of performance across diverse and challenging real-world scenarios (e.g., varying lighting, occlusions, non-standard plates).
    \item \textbf{Holistic System Integration:} The seamless integration of multi-object detection, specialized OCR, real-time feedback, and user-friendly interfaces into a single, deployable system is not consistently addressed.
\end{itemize}

Our work aims to address these limitations by proposing a holistic system. We specifically introduce the detection of rear-view mirror absence as a novel component in automated traffic safety. Furthermore, by employing YOLOv8, we aim for high accuracy and real-time performance (targeting $\sim$25 FPS, comparable or exceeding some existing systems \cite{Aquitan2024}), and by integrating a dedicated OCR enhancement pipeline, we target improved license plate recognition accuracy over systems that use generic OCR approaches. The selection of YOLOv8 is predicated on its documented advantages in speed and accuracy over predecessors like YOLOv5 and other architectures (as further discussed in Section \ref{sec:results_discussion}). This comprehensive approach, combining multiple violation detections with robust OCR and a practical interface, extends beyond the scope of most previous studies.

\section{Methodology}
\label{sec:methodology}
\begin{figure}[!t]
\centering
 \includegraphics[width=\columnwidth]{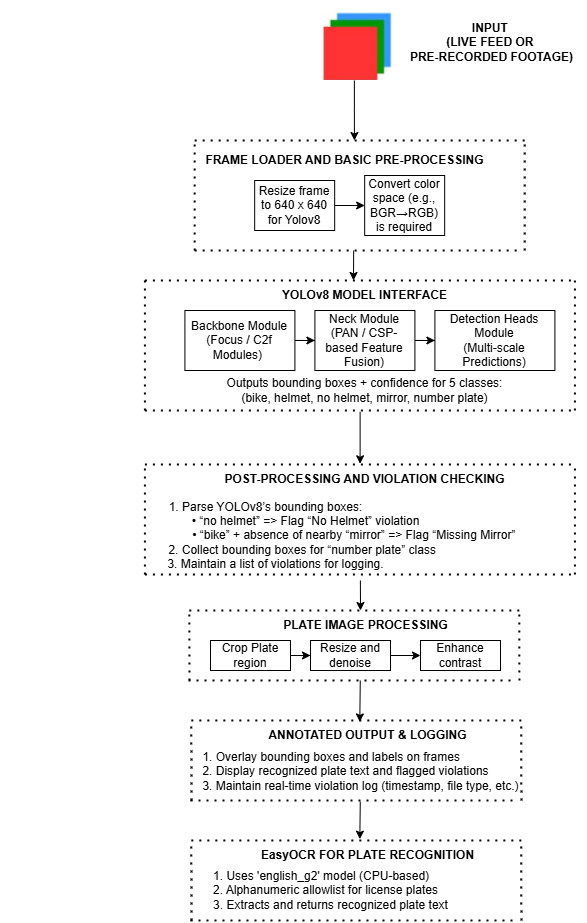} 
\caption{Block Diagram of the Proposed AI-Driven Traffic Violation Detection System. This illustrates the sequential flow from input processing and object detection with YOLOv8, through OCR-based license plate recognition, to violation logging and display via the Streamlit interface.}
\label{fig:block_diagram_methodology}
\end{figure}

The proposed system employs a multi-stage pipeline, integrating advanced object detection, optical character recognition (OCR), and a user-friendly interface to automate the identification of traffic violations. The core components of our methodology are detailed below (refer to Fig. \ref{fig:block_diagram_methodology} for a system overview).

\subsection{Model Selection and Object Detection}
The system employs YOLOv8 \cite{yolov8_ultralytics_ref} for real-time detection of five classes relevant to two-wheeler violations: bike, helmet, no helmet, mirror, and number plate. The model is initialized with COCO pre-trained weights \cite{coco_dataset_ref} and fine-tuned on the custom traffic dataset for 30 epochs using the Adam optimizer (lr=0.0001) with automatic mixed precision (AMP) to balance accuracy and training speed.

\subsection{Dataset Creation and Annotation}
\label{ssec:dataset}
A custom dataset comprising 466 diverse images was curated for training and validating the YOLOv8 model. These images were collected from various publicly available internet sources, capturing a range of real-world traffic scenarios, including different lighting conditions, vehicle densities, and camera perspectives, primarily focusing on Indian road contexts as indicated by license plate styles.

Each image was meticulously annotated using Roboflow \cite{roboflow_ref}, a platform facilitating efficient labeling and dataset management. The five object classes (`bike`, `helmet`, `no helmet`, `mirror`, `number plate`) were carefully bounded. To mitigate the relatively small base dataset size and enhance model generalization, extensive data augmentation techniques were applied via Roboflow. These included:
\begin{itemize}
    \item Geometric transformations: random rotations ($\pm$15 degrees), horizontal flips.
    \item Photometric distortions: brightness adjustments ($\pm$25\%), contrast changes, color jitter, and noise injection.
    \item Other augmentations like mosaic augmentation, which is inherent to YOLOv8 training pipelines.
\end{itemize}
These augmentations effectively increased the diversity and volume of training samples. The dataset was split into training (70\%), validation (20\%), and testing (10\%) sets to ensure robust model evaluation.

\subsection{Image Preprocessing for OCR}
\label{ssec:preprocessing_ocr}
Once a `number plate` is detected by YOLOv8, the corresponding region of interest (ROI) is cropped from the frame. To maximize the accuracy of the subsequent OCR step, this ROI undergoes a series of preprocessing techniques:
\begin{enumerate}
    \item \textbf{Grayscale Conversion:} The cropped license plate image is converted to grayscale, as color information is generally not pertinent for character recognition and can add unnecessary complexity.
    \item \textbf{Noise Reduction:} Bilateral filtering is applied. This technique is effective at smoothing images while preserving edges, which is crucial for maintaining the sharpness of characters on the license plate.
    \item \textbf{Contrast Enhancement:} Adaptive Histogram Equalization (AHE), specifically Contrast Limited Adaptive Histogram Equalization (CLAHE), is used. Unlike global histogram equalization, CLAHE operates on small regions (tiles) in the image, enhancing local contrast and making characters more distinct, especially under non-uniform illumination.
    \item \textbf{Binarization and Morphological Operations (Optional):} Depending on the image quality, adaptive thresholding might be applied to convert the image to binary, followed by morphological operations (e.g., opening or closing) to remove small noise artifacts or connect broken character segments. These steps are conditionally applied to avoid over-processing clean images.
\end{enumerate}
This tailored preprocessing pipeline significantly improves the quality of the input for the OCR engine, addressing challenges like poor lighting, minor blurs, and low resolution often encountered in real-world imagery. An ablation study quantifying the impact of these preprocessing steps is presented in Section \ref{sec:results_discussion}.

\subsection{Violation Detection Logic}
\label{ssec:violation_logic}
The system identifies traffic violations by analyzing the spatial relationships and presence (or absence) of detected objects within each frame:
\begin{enumerate}
    \item \textbf{Helmet Non-Compliance:} For each detected `bike`, the system searches for an associated `helmet` object within a predefined proximity to the rider(s) on the bike (approximated by the upper region of the `bike` bounding box). If a `bike` is detected but no `helmet` is found in the expected region, or if a `no helmet` class is directly detected in association with the `bike`, a helmet violation is flagged.
    \item \textbf{Rear-View Mirror Absence:} For each detected `bike`, the system checks for the presence of `mirror` objects. The expected locations for mirrors are typically on the handlebars, flanking the main body of the motorcycle. If a `bike` is detected and fewer than two `mirror` objects (or a user-configurable minimum, e.g., at least one for one side reflecting common legal requirements for at least one functional mirror) are found within a reasonable proximity and appropriate relative position to the `bike`'s bounding box, a missing mirror violation is flagged. The proximity threshold is empirically determined to accommodate variations in motorcycle designs and camera angles.
    \item \textbf{License Plate Association:} If a violation is flagged, the system attempts to associate it with a detected `number plate` that is spatially close to the violating `bike`.
\end{enumerate}
The use of proximity matching, based on Intersection over Union (IoU) or center point distances between relevant bounding boxes, ensures that violations are correctly attributed to the specific motorcycle involved. The confidence threshold for object detection, which influences violation flagging, was optimized using the F1-confidence curve analysis (detailed in Section \ref{sec:results_discussion}). 

\subsection{License Plate Recognition (LPR)}
\label{ssec:lpr}
License plate recognition is performed using EasyOCR \cite{easyocr_ref}, a Python library supporting multiple languages and known for its ease of use and good accuracy with minimal configuration. After a license plate ROI is detected and preprocessed (as described in Section \ref{ssec:preprocessing_ocr}), it is fed to the EasyOCR engine. We configure EasyOCR to primarily recognize English characters and digits, common in Indian license plates, using its `english g2` model which is optimized for GPU execution if available, otherwise defaulting to CPU. The raw text output from EasyOCR is then post-processed:
\begin{itemize}
    \item \textbf{Character Filtering:} An allowlist of alphanumeric characters typical for license plates is used to filter out noise or misrecognized symbols.
    \item \textbf{Formatting:} The recognized text is structured to match common license plate formats, potentially involving heuristics to correct common OCR errors (e.g., 'O' vs '0', 'I' vs '1').
\end{itemize}
This step ensures that the extracted registration numbers are as accurate and standardized as possible for logging and potential database queries.

\subsection{Frontend Development and Data Logging}
\label{ssec:frontend_logging}
To provide an intuitive user experience and facilitate real-time monitoring, a frontend interface is developed using Streamlit \cite{streamlit_ref}. This web application framework allows for rapid development of interactive UIs directly in Python. The interface:
\begin{itemize}
    \item Displays the live video feed (or processed video frames) annotated with bounding boxes for detected objects and violation flags.
    \item Shows the recognized license plate text alongside any flagged violations for a particular vehicle.
    \item Logs details of each detected violation, including timestamp, violation type (no helmet, missing mirror), the recognized license plate number, and a snapshot image of the violation, into a structured format (e.g., CSV file and image folder).
\end{itemize}
This user-friendly design enables efficient real-time monitoring and provides a persistent record for subsequent analysis or enforcement actions. The system processes input from live video feeds or pre-recorded footage frame by frame. The hardware specifications used for achieving the reported performance metrics are detailed in Section \ref{sec:results_discussion}.

\section{Results and Discussion}
\label{sec:results_discussion}
The proposed AI-system's performance in detecting traffic violations (helmet non-compliance, rear-view mirror absence) and recognizing license plates was rigorously evaluated. This section details quantitative metrics, model training dynamics, qualitative visual analysis, and real-time processing capabilities. Experiments were conducted on a system with an Intel Iris Xe GPU, Intel Core i5-1135G7 CPU, and 16GB RAM.

\subsection{Overall Detection Performance}
Quantitative results on our custom test set are in Table \ref{tab:yolov8_performance}. The system achieved an overall precision of 0.9147, recall of 0.886, and mAP@.50 of 0.843. The mAP@.50-.95 of 0.503 indicates robustness under stricter IoU thresholds. `Number Plate` detection was exceptional (Precision 0.981, mAP@.50 0.973). `No Helmet` detection was also strong. `Mirror` detection showed relatively lower metrics (Precision 0.729), attributable to their smaller size, variability, and occlusion potential.

\begin{table}[!htb]
\centering
\caption{YOLOv8 Model Performance Metrics.}
\label{tab:yolov8_performance}
\resizebox{\columnwidth}{!}{%
\begin{tabular}{@{}lcccc@{}}
\toprule
\textbf{Class} & \textbf{Precision} & \textbf{Recall} & \textbf{mAP@.50} & \textbf{mAP@.50-.95} \\ \midrule
Bike         & 0.876              & 0.832           & 0.861            & 0.517                \\
Helmet       & 0.870              & 0.845           & 0.872            & 0.514                \\
Mirror       & 0.729              & 0.715           & 0.733            & 0.508                \\
No Helmet    & 0.883              & 0.901           & 0.893            & 0.512                \\
Number Plate & 0.981              & 0.955           & 0.973            & 0.579                \\ \midrule
\textbf{All Classes (Overall)} & \textbf{0.9147}    & \textbf{0.886}  & \textbf{0.843}   & \textbf{0.503}       \\ \bottomrule
\end{tabular}%
}
\end{table}

\subsection{Model Training Dynamics and Analysis}
\begin{figure*}[!t] 
\centering
 \includegraphics[width=0.95\textwidth]{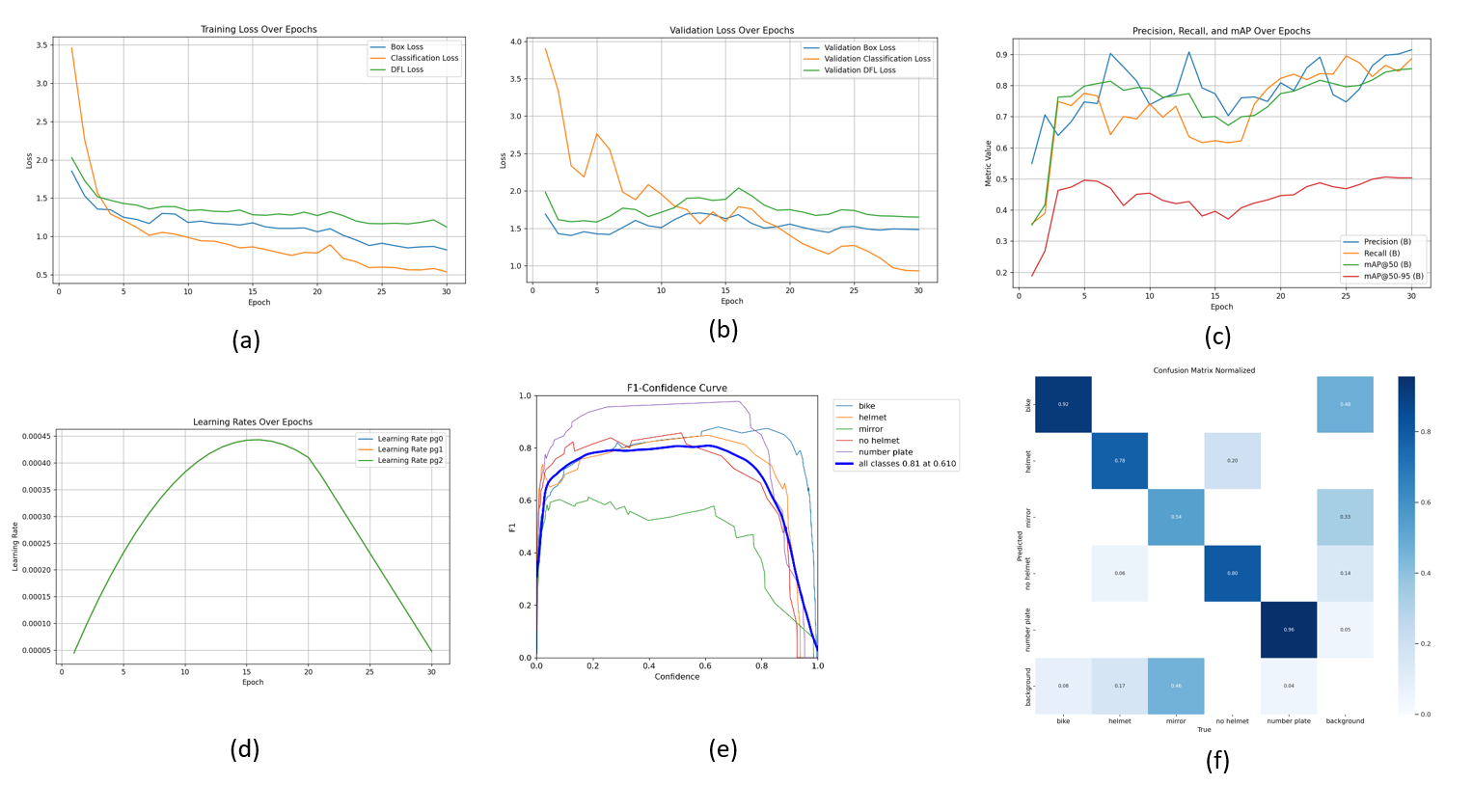} 
\caption{Model Training Dynamics: (a) Training Loss. (b) Validation Loss. (c) P, R, \& mAP. (d) Learning Rate. (e) F1-Confidence. (f) Confusion Matrix.}
\label{fig:training_dynamics_composite}
\end{figure*}

Fig. \ref{fig:training_dynamics_composite} presents a consolidated view of the model's training dynamics over 30 epochs. Training losses (Fig. \ref{fig:training_dynamics_composite}a) show steady decreases. Validation losses (Fig. \ref{fig:training_dynamics_composite}b) converge similarly, suggesting good generalization without significant overfitting. Precision, recall, and mAP@.50 (Fig. \ref{fig:training_dynamics_composite}c) align with Table \ref{tab:yolov8_performance}. The learning rate schedule (Fig. \ref{fig:training_dynamics_composite}d) was well-calibrated. An optimal detection confidence threshold of 0.610 (F1-score 0.81) was derived from the F1-Confidence Curve (Fig. \ref{fig:training_dynamics_composite}e). The normalized confusion matrix (Fig. \ref{fig:training_dynamics_composite}f) confirms strong classification accuracy and balanced performance post-augmentation.

\subsection{Qualitative Analysis and OCR Performance}
\begin{figure}[!htb] 
\centering
 \includegraphics[width=0.9\columnwidth]{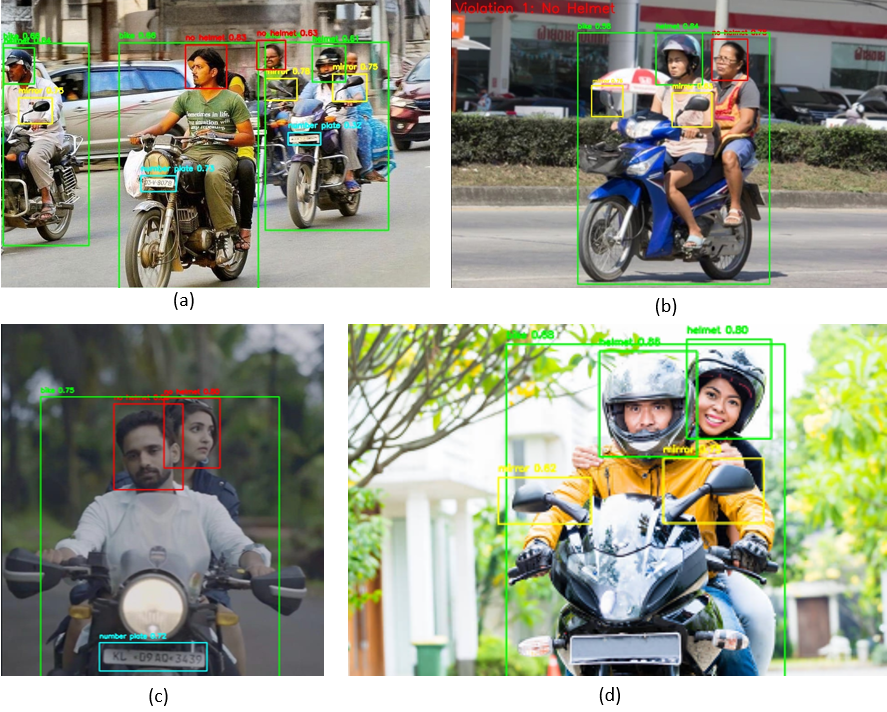} 
\caption{System output examples: (a) Multiple violators. (b) Rider/pillion non-compliance. (c) LP detection \& OCR. (d) No violations.}
\label{fig:detection_examples}
\end{figure}

\begin{figure}[!htb]
\centering
 \includegraphics[width=0.6\columnwidth]{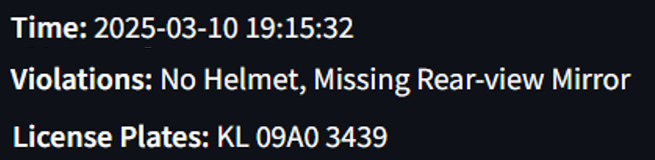} 
\caption{Streamlit interface violation log.}
\label{fig:streamlit_interface_log}
\end{figure}

Fig. \ref{fig:detection_examples} illustrates the system's output in diverse scenarios: (a) multiple violators in a dense scene; (b) helmet non-compliance for rider and pillion; (c) successful license plate detection (conf: 0.72) and OCR ("KL 09A0 3439") despite an oblique angle, with `bike` (conf: 0.75) and `no helmet` (conf: 0.81, 0.83) also identified; (d) a compliant scenario with no violations detected. The overall 85-87\% OCR success rate is notable, though challenges in severe low light/blur persist. The Streamlit interface (Fig. \ref{fig:streamlit_interface_log}) logs these violations.

\subsection{Ablation Study on OCR Preprocessing}
The impact of OCR preprocessing (Sec. \ref{ssec:preprocessing_ocr}) was quantified on a challenging 100-image subset (Table \ref{tab:ocr_ablation}). The full pipeline (Grayscale + Bilateral Filter + CLAHE) improved OCR accuracy from 62\% (raw) to 87\%, validating its necessity.

\begin{table}[!htb]
\centering
\caption{Ablation Study: OCR Accuracy (\%).}
\label{tab:ocr_ablation}
\resizebox{0.85\columnwidth}{!}{%
\begin{tabular}{@{}lc@{}}
\toprule
\textbf{Preprocessing Applied}      & \textbf{OCR Accuracy (\%)} \\ \midrule
None (Raw Crop)                 & 62                       \\
Grayscale only                  & 68                       \\
Grayscale + Bilateral Filter    & 75                       \\
Grayscale + Bilateral + CLAHE   & 87                       \\ \bottomrule
\end{tabular}%
}
\end{table}

\subsection{Real-time Processing and Comparative Analysis}
The system processes $\sim$25 FPS on the specified hardware, suitable for real-time monitoring. Table \ref{tab:model_comparison} compares our custom YOLOv8 system with others from literature \cite{Bao2023improvedYOLOv8}. Our system shows competitive or superior performance, especially in precision and mAP@.50 for our multi-class task.

\begin{table}[!htb]
\centering
\caption{Comparison with Other Models (Adapted from \cite{Bao2023improvedYOLOv8} and internal tests).}
\label{tab:model_comparison}
\resizebox{\columnwidth}{!}{%
\begin{tabular}{@{}lcccc@{}}
\toprule
\textbf{Model}    & \textbf{Prec.} & \textbf{Recall} & \textbf{mAP@.50} & \textbf{mAP@.50-.95} \\ \midrule
YOLOv3 (gen.)  & $\sim$0.60         & $\sim$0.50      & $\sim$0.55       & $\sim$0.33           \\
YOLOv5m (gen.) & $\sim$0.70         & $\sim$0.74      & $\sim$0.74       & $\sim$0.51           \\
RPN \cite{Bao2023improvedYOLOv8} & 0.597       & 0.894        & 0.779         & -                  \\
Fast-RCNN \cite{Bao2023improvedYOLOv8} & 0.672    & 0.804        & 0.774         & -                  \\
\textbf{Our System} & \textbf{0.9147} & \textbf{0.886} & \textbf{0.843} & \textbf{0.503}    \\ \bottomrule
\end{tabular}%
}
\end{table}

\subsection{Discussion of Limitations}
Limitations include dataset size/geographic focus, potential performance degradation in severe adverse conditions (not extensively tested), and need for optimization for edge deployment. Mirror detection can be affected by occlusion. Adversarial attacks and detailed ethical/privacy LPR data considerations are key future work.

\section{Conclusion}
\label{sec:conclusion}
This paper presented a novel AI-driven system for real-time detection of multiple critical traffic violations on two-wheelers, specifically helmet non-compliance, the absence of rear-view mirrors, and license plate recognition using YOLOv8 and EasyOCR. The system demonstrated high overall accuracy (mAP@.50 of 0.843, Precision 0.915, Recall 0.886) and achieved promising violation-specific detection rates, notably pioneering automated rear-view mirror checks. Key contributions include the holistic integration of these detection tasks, an effective OCR preprocessing pipeline enhancing license plate readability (85-87\% success), and a user-friendly Streamlit interface for real-time monitoring and logging, validated at $\sim$25 FPS.

While the system performs robustly, the relatively lower performance for mirror detection (Precision 0.729) highlights an area for further refinement. The current work successfully establishes a scalable and efficient framework for automating traffic rule enforcement. By significantly reducing the burden on manual inspection and improving the consistency of enforcement, this research offers a practical pathway towards enhancing road safety and promoting adherence to traffic regulations. Future efforts will focus on expanding dataset diversity, improving model robustness in challenging conditions, and exploring on-device deployment.

\section{Future Work}
\label{sec:future_work}
Future work includes enhancing system capabilities and real-world applicability:
\begin{itemize}
    \item \textbf{Dataset Expansion and Diversity:} Expanding the dataset with diverse scenarios (geographic, lighting, weather, non-standard plates) and incorporating advanced image analysis techniques for better object understanding \cite{pa2023static} to improve generalization and OCR accuracy.
    \item \textbf{Model Robustness and Performance:} Further optimizing model performance, especially for challenging detections like rear-view mirrors, by exploring advanced semantic segmentation \cite{karanth2024saree}, temporal analysis, and newer AI architectures (e.g., transformers).
    \item \textbf{Edge Deployment and Efficiency:} Adapting the system for resource-constrained edge devices (e.g., NVIDIA Jetson) using model quantization and pruning for low-latency, decentralized processing, building on real-time system frameworks \cite{raman2020hybrid}.
    \item \textbf{OCR Enhancements:} Improving OCR for distorted/low-resolution plates and adding multilingual support.
    \item \textbf{System Integration and Ethical Considerations:} Exploring secure integration with official databases and broader smart city frameworks (e.g., cloud-based smart parking solutions \cite{pandit2019cloud}), while rigorously addressing data privacy and ethical implications.
\end{itemize}
These advancements aim for a more comprehensive, robust, and ethically responsible traffic monitoring solution.

\end{document}